\documentclass[12pt]{article}
\usepackage[dvips]{color,graphicx}
\usepackage{amsmath,amssymb}
\usepackage{epsf}
\usepackage{natbib}
\usepackage{subcaption}

\newtheorem{theorem}{Theorem}
\newtheorem{lemma}[theorem]{Lemma}

\newtheorem{example}[theorem]{Example}

\title{Accuracy of Latent-Variable Estimation\\
in Bayesian Semi-Supervised Learning 
\thanks{This research was partially supported by the Kayamori Foundation of Informational Science Advancement
and KAKENHI 23500172 and 24700139.}
}
\author{Keisuke Yamazaki\\
       k-yam@math.dis.titech.ac.jp \\
       Department of Computational Intelligence and Systems Science,\\
       Tokyo Institute of Technology\\
       G5-19 4259 Nagatsuta Midori-ku Yokohama, Japan
	}
\date{}
\begin{document}
\sloppy
\maketitle

\begin{abstract}
Hierarchical probabilistic models, such as Gaussian mixture models,
are widely used for unsupervised learning tasks.
These models consist of observable and latent variables,
which represent the observable data and the underlying data-generation process, respectively.
Unsupervised learning tasks, such as cluster analysis, are regarded as estimations of
latent variables based on the observable ones.
The estimation of latent variables in semi-supervised learning, where some labels are observed,
will be more precise than that in unsupervised,
and one of the concerns is to clarify the effect of the labeled data.
However, there has not been sufficient theoretical analysis of the accuracy of the estimation of latent variables.
In a previous study, a distribution-based error function was formulated,
and its asymptotic form was calculated for unsupervised learning with generative models.
It has been shown that, for the estimation of latent variables, the Bayes method is more accurate than the maximum-likelihood method.
The present paper reveals the asymptotic forms of the error function
in Bayesian semi-supervised learning for both discriminative and generative models.
The results show that the generative model, which uses all of the given data,
performs better when the model is well specified.
\\
{\it keywords:} Latent-variable estimation, Generative and discriminative models,
Bayes statistics
\end{abstract}

\section{Introduction}
\sloppy
Hierarchical statistical models, such as the Gaussian mixture model, 
are widely employed in unsupervised learning.
They consist of observable and latent variables,
which express the given data and the underlying data-generation process, respectively.
A typical task of unsupervised learning is clustering,
in which observable data is used to estimate labels that indicate which cluster the data are from.
The Gaussian mixture model is often used for cluster analysis, and in this model,
a Gaussian component represents a cluster.
If the parameter is known,
the probabilities of the labels for each data point are easily computed.
However, in many practical situations, the parameter is unknown,
and both it and the latent variable must be estimated.
Some learning algorithms, such as the expectation--maximization (EM) algorithm
\citep{Dempster77}
and the variational Bayes method \citep{Attias99,Ghahramani00,Beal03},
have two explicit estimation steps (the E step and the M step).
The present paper focuses on the Bayesian approach,
which uses the posterior distribution to marginalize out the parameter
and calculates the probabilities of the latent variables.

There are two ways to use hierarchical models:
to predict unseen observable data or to estimate the latent variables.
The accuracy of prediction has been analyzed theoretically.
The generalization error measures the accuracy, 
and, in many cases, its asymptotic behavior is well known.
When the error function is defined by the Kullback-Leibler divergence,
the asymptotic form of the error is well known in the maximum-likelihood method,
and it has been used as a criterion for selecting model complexity \citep{Akaike,Takeuchi76,Econometrica:White:1982}.
In the Bayes method, the posterior distribution of the parameter plays an important role
to determine the accuracy;
the normalizing factor of the distribution is the marginal likelihood
and its negative logarithm has a direct relation with the error function \citep{Levin}.
Since the asymptotic form of the marginal likelihood has been calculated \citep{Schwarz,Clarke90},
this relation allows us to derive the asymptotic generalization error.

On the other hand, latent-variable (LV) estimation has not been analyzed sufficiently.
Recently, a distribution-based error function was formulated to determine the accuracy of the LV estimation, 
and its asymptotic form was calculated \citep{Yamazaki14a,Yamazaki15a}.
The results showed the different properties from those of the estimation of observable variables (OVs).
For example, the Bayes LV estimation is more accurate than the maximum-likelihood estimation
under the regularity conditions while these estimation methods have the same accuracy in the OV prediction.
Moreover, when there are unnecessary labels in the model,
the Bayes method automatically eliminates the redundant labels.
Because of these advantages, we focus on the Bayes approach in the present paper.

There are discriminative and generative approaches to defining the model;
the discriminative approach results in a model that expresses the causal effect of the observable data
on the latent variables,
while the generative approach results in a model that explains the data-generation process.
Our previous study mainly analyzed the generative model \citep{Yamazaki14a}.
The present paper compares these approaches in terms of the estimation of LVs.

In the estimation of LVs,
partially observed labels will improve accuracy
because they have information about the targets of the estimation.
Learning from a mixture of labeled and unlabeled data is referred to as
semi-supervised learning \citep{Zhu:Survey,Yamazaki15b}.
Its main concerns include clarifying how the supplemental information affects the accuracy of the estimation
and developing a learning algorithm that achieves better results based on this advantage.

The present paper analyzes the statistical properties of the Bayes method for semi-supervised learning,
and the main contributions are as follows:
\begin{enumerate}
\item The asymptotic forms of the error function are derived for both the discriminative and generative models.
\item The generative model asymptotically performs better in the well-specified case.
\end{enumerate}
Asymptotic analysis generally assumes that the amount of unlabeled data is sufficiently large.
If the number of labels that are given is not large, there may be 
no effect on the estimation, or it may be very weak.
To magnify the effect of the observed labels, we assume that the number of labeled data points is $\alpha n$,
where $n$ is the number of the training data points and $\alpha$ is the ratio of the labels
where $0<\alpha<1$ (see Fig.~\ref{fig:data_structure}).

The rest of this paper is organized as follows.
The next section formalizes the data structure and the model expression.
In Section \ref{sec:bayes_errorfunc}, we introduce 
the Bayesian LV estimation and an error function to measure its accuracy.
The discriminative and the generative approaches are explained.
Section \ref{sec:analysis} shows the asymptotic analysis of the error function
and compares the approaches.
Section \ref{sec:dis}
discusses the magnitude relation of the error function among the approaches
and clarifies the effect of the observable labels on the accuracy.
%
\section{Data Structure and Model Expression}
\label{sec:data_model}
\subsection{Formal Expressions of Data and Model}
This subsection formulates the settings of the data and the model.

\begin{figure}[t]
\begin{center}
\includegraphics[angle=-90,width=0.8\columnwidth]{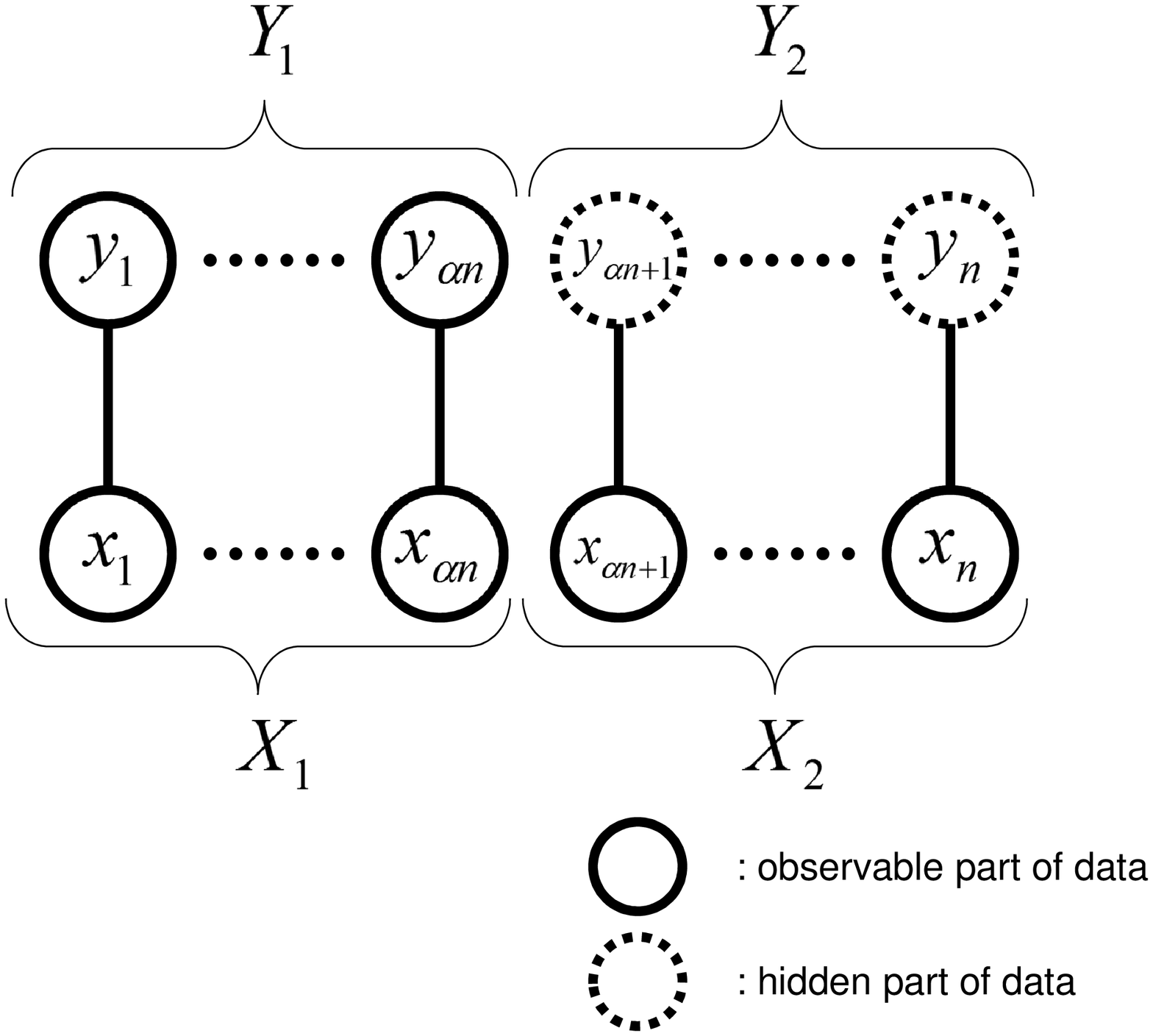}
\caption{Data structure: pairs $(x_i,y_i)$ are generated. The solid circles
represent observable data, which is used for training, and the dashed circles represent hidden data,
which are the target of the estimation.}
\label{fig:data_structure}
\end{center}
\end{figure}
Fig.~\ref{fig:data_structure} shows the structure of the data;
the observable and latent variable are $x\in R^M$ and $y\in \{1,\dots,K\}$, respectively.
The data points $\{(x_1,y_1),\dots,(x_n,y_n)\}$ are independent and identically distributed,
and $\alpha n$ data points $\{(x_1,y_1),\dots,(x_{\alpha n},y_{\alpha n})\}$ are labeled,
where $0<\alpha<1$ and $\alpha n$ is an integer.
We define the following data sets:
\begin{align*}
X_1 &= \{x_1, \dots, x_{\alpha n}\},\\
X_2 &= \{ x_{\alpha n+1}, \dots, x_n\},\\
Y_1 &= \{ y_1, \dots, y_{\alpha n} \},\\
Y_2 &= \{ y_{\alpha n +1}, \dots, y_n\},\\
X^n &= \{X_1,X_2\},\\
Y^n &= \{Y_1,Y_2\},\\
D &= \{ X_1, Y_1, X_2 \} = \{X^n, Y_1\},
\end{align*}
where $X_1$ is the set of $\alpha n$ observable data points,
$Y_1$ is the set of the corresponding labels,
and the target of the LV estimation is $Y_2$.
The set $D$ contains the available data for the estimation.
The number of the labels grows linearly with the total number of data points $n$.

The generative model represents the underlying process of data generation.
In the present paper, we assume that the observable variables are caused by the latent variables.
The mathematical expression for this is $p(x,y|w)=p(x|y,w)p(y|w)$,
where $w$ is the parameter.
On the other hand,
the discriminative model expresses the probability of the latent variable
based on the observable variable;
the classifier of the learning model is defined by $p(y|x,w)$.
If the discriminative model is defined on the basis of the generative one,
the model expression is given by $p(y|x,w)=p(x,y|w)/p(x|w)$, where $p(x|w)=\sum_y p(x,y|w)$.
According to Fig.~\ref{fig:data_structure},
the target of the LV estimation is $p(Y_2|D)$.
There are various ways to define $p(Y_2|D)$, as will be shown in the next section.
Let $q(y|x)$ and $q(x)$ be the true classifier and the true density of $x$, respectively;
the data $(X^n,Y^n)$ are generated from $q(x,y)=q(y|x)q(x)$.

\subsection{An Example of the Model}
For illustrative purposes,
we present the following data source and model:
\begin{example}[Data distribution]
\label{ex:data}
Define $x$ and $y$ so that $x\in R$ and $y\in \{1,2\}$.
The sample data follow the following distribution:
\begin{align}
q(x,y) = a^*_y \mathcal{N}(x|b^*_y,\sigma),
\end{align}
where $a^*_y$ is the mixing ratio and $\mathcal{N}(x|\mu,\sigma)$
is the one-dimensional Gaussian distribution with mean $\mu\in R$
and variance $\sigma\in R_{>0}$.
The mixing ratio is expressed as $a^*_1=a^*\in (0,1)$ and $a^*_2=1-a^*$.
\end{example}
Note that the density of $x$ is also based on $q(x,y)$:
\begin{align*}
q(x) =& \sum_{y=1}^2 q(x,y).
\end{align*}
%
%
%
%
\begin{example}[Learning model]
\label{ex:model}
The two-component Gaussian mixture learning model is defined by
\begin{align*}
p(x|w) =& \sum_{k=1}^2 a_k \mathcal{N}(x|b_k,\sigma),
\end{align*}
where $0\le a_1 \le 1$, $a_2 = 1-a_1$, and $b_k\in R$.
The parameter consists of
$w=(a_1, b_1, b_2)^\top$.
The discriminative model is based on the following mixture:
\begin{align}
p(y=k|x,w) =& \frac{p(x,y=k,w)}{\sum_{j=1}^2 p(x,y=j,w)} \nonumber\\
=& \frac{a_k\mathcal{N}(x|b_k,\sigma)}{\sum_{j=1}^2a_j\mathcal{N}(x|b_j,\sigma)}. \label{eq:def_exmodel}
\end{align}
\end{example}
There exists a true parameter $w^*$ such that $q(y|x)=p(y|x,w^*)$.
\begin{figure}[t]
\centering
\includegraphics[width=0.7\linewidth]{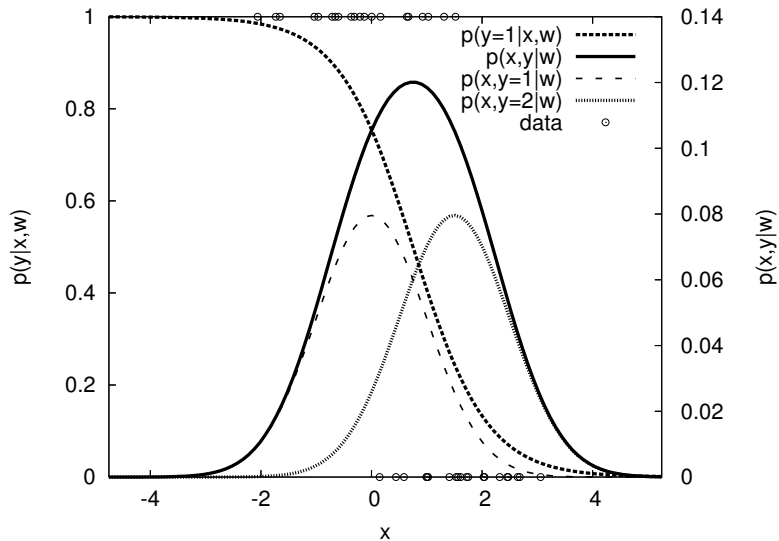}
\caption{Model shapes with the generative and the discriminative expressions.}
\label{fig:gd_models}
\end{figure}
Fig.~\ref{fig:gd_models} shows the model shapes with the generative and the discriminative expressions
in the two-component Gaussian mixture.
The true parameter is $w^*=(a_1^*,b^*_1,b^*_2)^\top=(0.5,0,1.5)^\top$.
A sample set of data from $q(x,y)$ is shown in the same figure;
the data with $y=1$ and $y=2$ are on the upper and the lower horizontal lines, respectively.

\subsection{Redundant Parameters in the Discriminative Model}
\label{sec:redundancy}
The classifier $p(y|x,w)$ based on $p(x,y|w)$ does not provide a one-to-one relation
between the model expression and the parameter.
Let us consider the case in Examples \ref{ex:data} and \ref{ex:model}. 
Suppose that $a^*_k\neq 0$, which means that the model has the same number of clusters as the data distribution.
Even though there is no redundancy in the number of clusters,
there is redundancy in the parameter:
\begin{align*}
p(y=1|x,w) =& \frac{a_1\exp\{-(x-b_1)^2/(2\sigma^2)\}}
{Z_m}\\
=& \frac{1}{1+f_1(x,w)},
\end{align*}
where the normalizing factor is expressed as
\begin{align*}
Z_m =& \sum_{y=1}^2 a_y\exp\{-(x-b_y)^2/(2\sigma^2)\},
\end{align*}
and the functions $f_1$ and $f_2$ are written as
\begin{align*}
\ln f_1(x,w) =& \frac{1}{\sigma^2}(b_2-b_1)x
-\frac{1}{2}(b_2^2-b_1^2) +\ln\frac{a_2}{a_1}.
\end{align*}
The coefficients of $x$
and the constant terms contain more elements of the parameter than are needed to express the function.
Let us reparameterize $f_1(x,w)$ as
\begin{align*}
\ln f_1(x,\bar{w}) =& c_1x+c_2,
\end{align*}
where $\bar{w}=(c_1,c_2)^\top$.
Considering the case $y=2$, we can easily confirm that the parameter $\bar{w}$
is sufficient to express $p(y=2|x,w)$.
This means that the essential dimension of the parameter is $\dim\bar{w}=2$
instead of $\dim w=3$.
To eliminate the redundancy, we regard $a_k$ as a positive constant
and let the reduced parameter be $\bar{w}$,
which consists of the elements of $w$ except for $a_k$.
For the general dimension of data $M$
and the general number of the components $K$, 
we can calculate that $\dim\bar{w}=\dim w-M$ in the Gaussian mixture.
\section{The Bayes LV Estimation and Error Function}
\label{sec:bayes_errorfunc}
This section introduces the Bayes LV estimation
and an error function to measure its accuracy.
Let $L(w,X^n,Y^n)$ be a likelihood function on $\{X^n,Y^n\}$,
and let $L(w,D)=\sum_{Y_2}L(w,X^n,Y^n)$ be one on $D$.
In the Bayesian model, the LV estimation, which corresponds to constructing $p(Y_2|D)$,
is written as
\begin{align}
p(Y_2|D) = \frac{\int L(w,X^n,Y^n)\varphi(w|\eta)dw}{\int L(w,D)\varphi(w|\eta)dw},\label{eq:1stdef}
\end{align}
where $\varphi(w|\eta)$ is a prior distribution and $\eta$ is the hyperparameter.
We will consider here the following three likelihood functions:
\begin{description}
\item{(Model 1)}
\begin{align*}
L_1(\bar{w},X^n,Y^n) =& \prod_{i=1}^n p(y_i|x_i,\bar{w}),\\
L_1(\bar{w},D) =& \prod_{i=1}^{\alpha n} p(y_i|x_i,\bar{w});
\end{align*}
\item{(Model 2)}
\begin{align*}
L_2(w,X^n,Y^n) =& \prod_{i=1}^{\alpha n}p(y_i|x_i,w)\prod_{i=\alpha n+1}^n p(x_i,y_i|w),\\
L_2(w,D) =& \prod_{i=1}^{\alpha n}p(y_i|x_i,w)\prod_{i=\alpha n+1}^n p(x_i|w);
\end{align*}
\item{(Model 3)}
\begin{align*}
L_3(w,X^n,Y^n) =& \prod_{i=1}^n p(x_i,y_i|w),\\
L_3(w,D) =& \prod_{i=1}^{\alpha n}p(x_i,y_i|w)\prod_{i=\alpha n+1}^n p(x_i|w).
\end{align*}
\end{description}
The first and third models correspond to the discriminative and generative models, respectively.
The second one is a hybrid model; the labeled data are used in the discriminative expression,
and the unlabeled data are used in the generative expression.
Note that Model 2 cannot use the reduced parameterization $\bar{w}$ in the discriminative expression,
since it must use $w$ in the generative part.

The formulation defined by Eq.~\ref{eq:1stdef} has the following equivalent expressions:
\begin{align}
p(Y_2|D) =& \int \prod_{i=\alpha n+1}^n p(y_i|x_i,w) p(w|D)dw, \label{eq:2nddef}\\
p(w|D) =& \frac{L(w,D)\varphi(w|\eta)}{\int L(w,D)\varphi(w|\eta)dw}, \label{eq:posterior}
\end{align}
where $p(w|D)$ is the posterior distribution. 
The first definition of $p(Y_2|D)$, Eq.~\ref{eq:1stdef},
is used for theoretical calculations in the asymptotic analysis,
and the second one, Eq.~\ref{eq:2nddef}, is useful for numerical computations
and for comparison of the models.
Note that for the both definitions the parameter $w$ is replaced with $\bar{w}$ in Model 1.
\begin{figure}[t]
\begin{minipage}[b]{.3\linewidth}
\centering
\includegraphics[width=\linewidth]{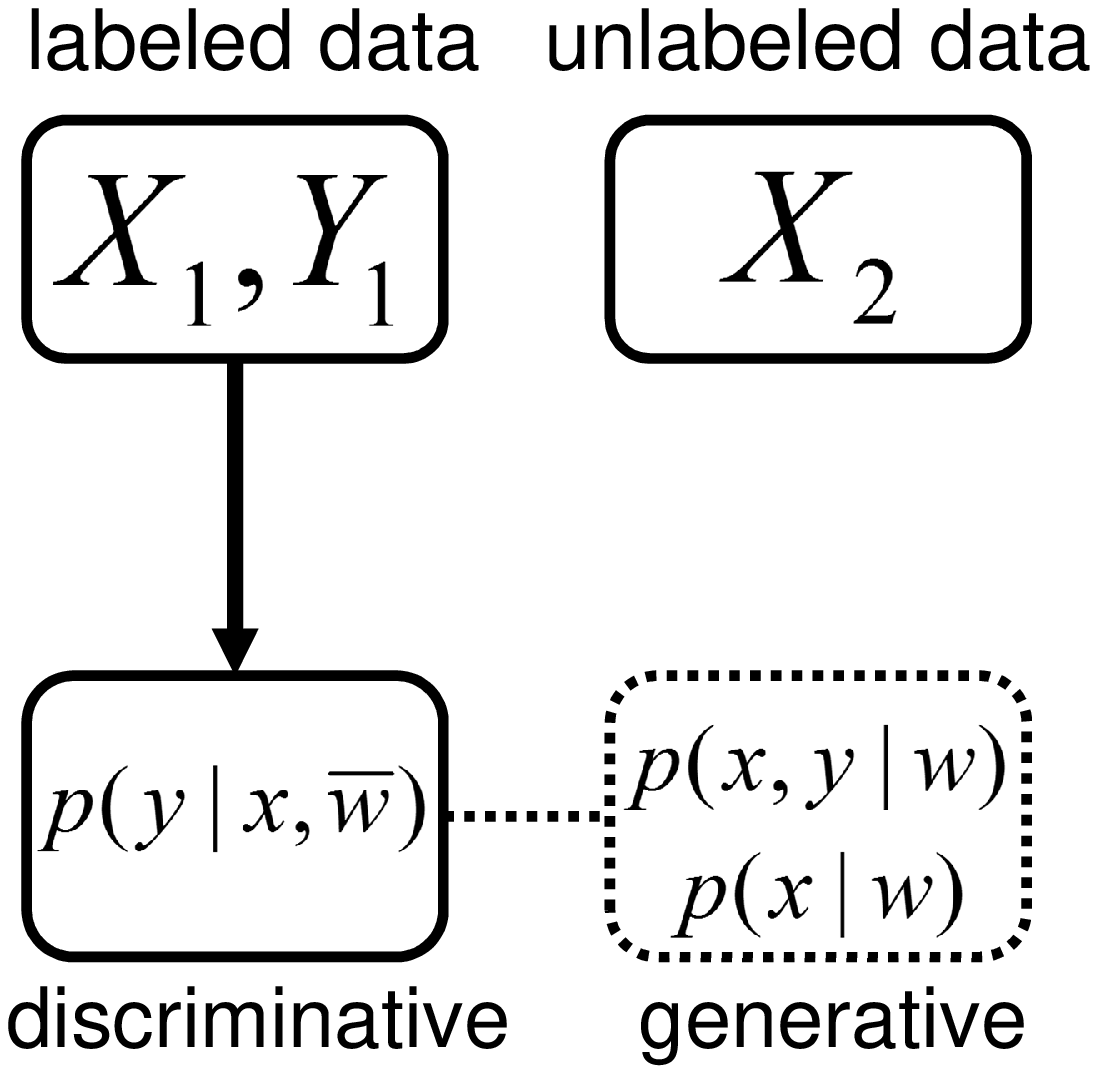}
\subcaption{Model 1}
\end{minipage}\;\;\;
\begin{minipage}[b]{.3\linewidth}
\centering
\includegraphics[width=\linewidth]{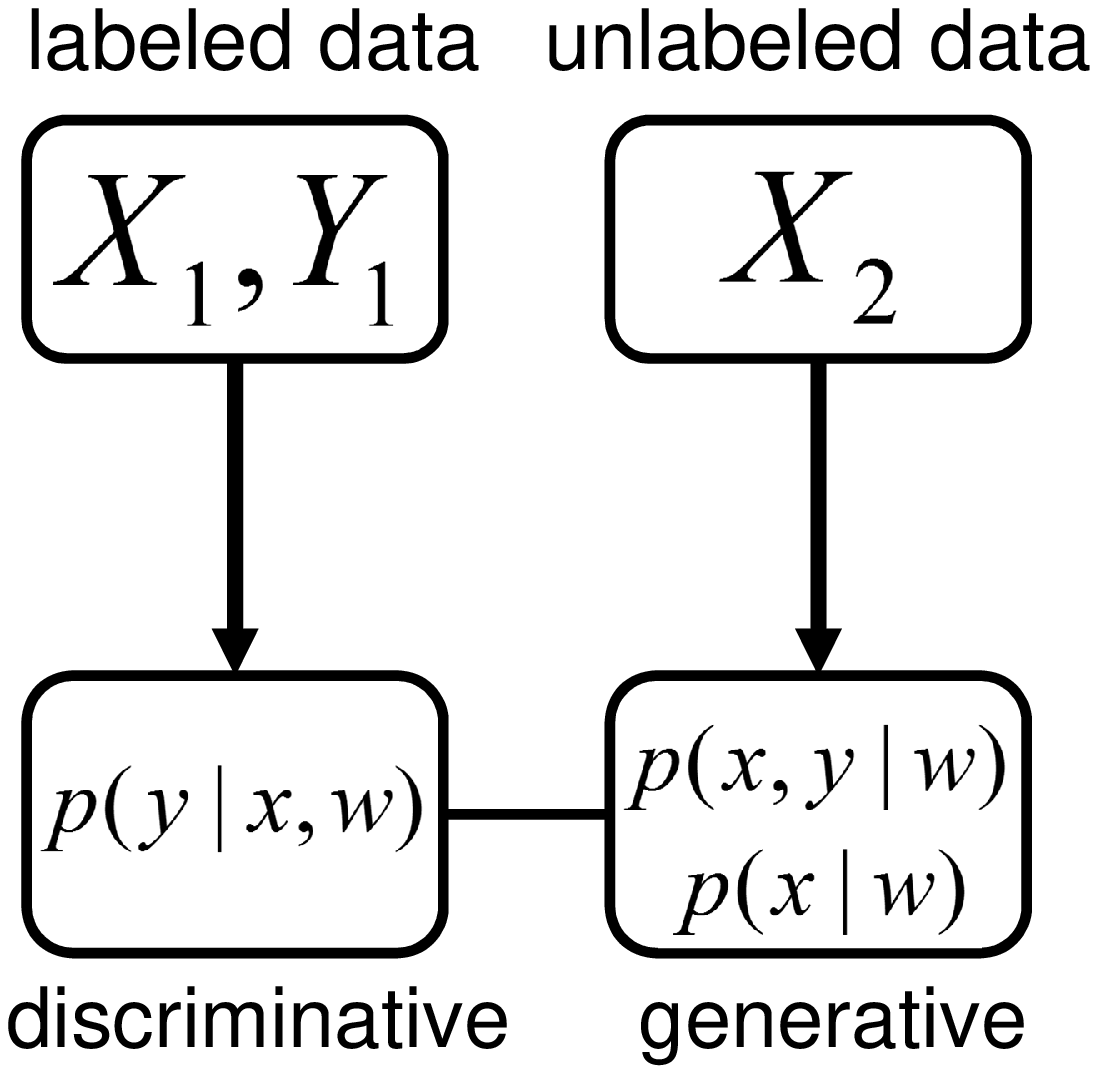}
\subcaption{Model 2}
\end{minipage}\;\;\;
\begin{minipage}[b]{.3\linewidth}
\centering
\includegraphics[width=\linewidth]{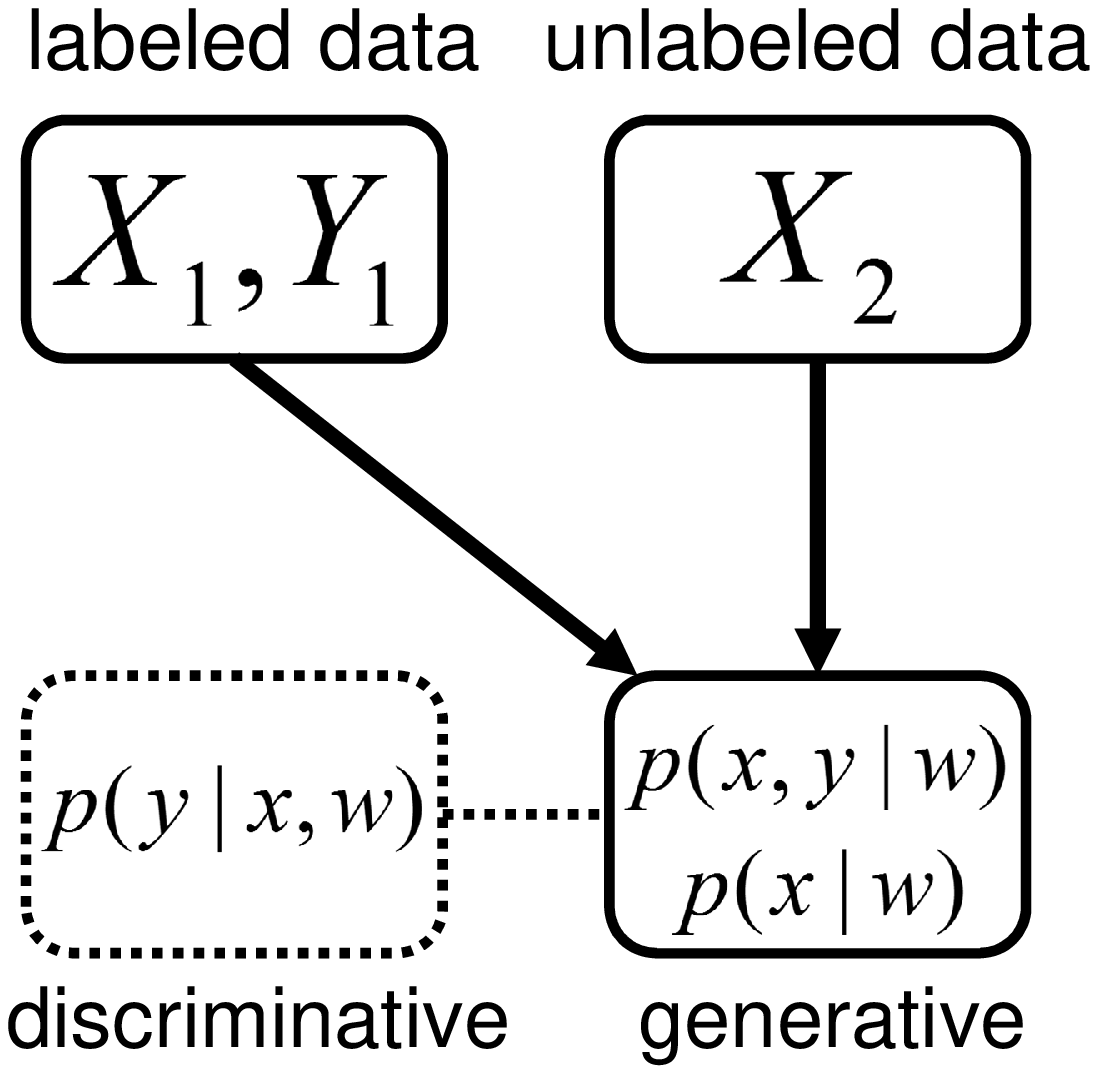}
\subcaption{Model 3}
\end{minipage}
\caption{The relations between the data sets and the model expressions}
\label{fig:rel_data_model}
\end{figure}
Fig.\ref{fig:rel_data_model} shows relations between the data sets and the model expressions
in the posterior distribution.
For example, in Model 2, the main factor of the posterior distribution $L(w,D)$
contains the discriminative expression with the labeled data $\{X_1,Y_1\}$
and the generative expression with the unlabeled data $X_2$.
Note that $L(\bar{w},D)$ in Model 1 contains the discriminative expression only,
where the unlabeled data $X_2$ are not used for the estimation.

Since the data are i.i.d., the true joint probability of $Y_2$ is given by
\begin{align}
q(Y_2|D) = q(Y_2|X_2) = \prod_{i=\alpha n+1}^n q(y_i|x_i). \label{eq:true}
\end{align}
The error function is defined as the difference between the true and the estimated
probabilities of $Y_2$.
Following previous work \citep{Yamazaki14a},
the error is based on the Kullback--Leibler divergence:
\begin{align*}
D(n) = \frac{1}{(1-\alpha)n}E_D\bigg[ \sum_{Y_2} q(Y_2|D)\ln\frac{q(Y_2|D)}{p(Y_2|D)}\bigg],
\end{align*}
where the expectation means
\begin{align*}
E_D[f(X_1,Y_1,X_2)] =& \int \sum_{Y_1} f(X_1,Y_1,X_2) q(Y_1|X_1)q(X_1)q(X_2)dX_1dX_2.
\end{align*}
Since the number of elements in $Y_2$ is $(1-\alpha)n$,
the error function is the average divergence for one latent variable.
\section{Asymptotic Analysis of the Error Function}
\label{sec:analysis}
This section shows one of the main results of this paper:
the asymptotic forms of the error functions of the three models
and a comparison between them.

We assume the following condition:
\begin{description}
\item[(A1)] In the discriminative expression,
there exists a true parameter $\bar{w}^*$ such that $q(y|x)=p(y|x,w^*)$
in the support of $\varphi(\bar{w}|\eta)$, and the following Fisher information matrix
in the neighborhood of $\bar{w}^*$ exists
and is positive definite:
\begin{align*}
\{ I_{y|x}(\bar{w})\}_{ij} =& \int \sum_y \frac{\partial \ln p(y|x,\bar{w})}{\partial \bar{w}_i}
\frac{\partial \ln p(y|x,\bar{w})}{\partial \bar{w}_j}p(y|x,\bar{w})q(x)dx.
\end{align*}
\item[(A2)] In the generative expression,
there exists a true parameter $w^*$ such that $q(x,y)=p(x,y|w^*)$
in the support of $\varphi(w|\eta)$, and the following Fisher information matrices 
in the neighborhood of $w^*$ exist
and are positive definite:
\begin{align*}
\{ I_{xy}(w)\}_{ij} =& \int \sum_y \frac{\partial \ln p(x,y|w)}{\partial w_i}
\frac{\partial \ln p(x,y|w)}{\partial w_j}p(x,y|w)dx,\\
\{ I_x(w)\}_{ij} =& \int \frac{\partial \ln p(x|w)}{\partial w_i}
\frac{\partial \ln p(x|w)}{\partial w_j}p(x|w)dx.
\end{align*}
\end{description}
These conditions indicate the ideal situation for the estimation,
where the estimated probability $p(Y_2|D)$ in all models converges to the true one
and the model is identifiable.

When the discriminative model $p(y|x,w)$ is based on the generative expression, such as in Example \ref{ex:model},
let another Fisher information matrix be defined by
\begin{align*}
\{ I(w)\}_{ij} =& \int \sum_y \frac{\partial \ln p(y|x,w)}{\partial w_i}
\frac{\partial \ln p(y|x,w)}{\partial w_j}p(y|x,w)q(x)dx,
\end{align*}
where $I(w^*) = I_{xy}(w^*)-I_x(w^*)$.
Note that here we use the common parameter setting and not the reduced one $\bar{w}$;
this means that, due to the redundancy of the parameters, the rank of $I(w^*)$ is not more than $\dim w-M$ in the Gaussian mixture.
The parameter redundancy in the discriminative model must be eliminated for the condition (A1).
Thus, we use the notation $\bar{w}$ to indicate the parameter setting satisfying (A1).

The following theorem shows the asymptotic behavior of the error function.
\begin{theorem}
\label{th:3asympError}
Let $D_1(n)$, $D_2(n)$, and $D_3(n)$ be the error functions of Models 1, 2, and 3, respectively.
Under the conditions (A1) and (A2), it holds that
\begin{align*}
D_1(n) =& \frac{\dim \bar{w}}{2}\frac{\ln 1/\alpha}{1-\alpha}\frac{1}{n} + o\bigg(\frac{1}{n}\bigg),\\
D_2(n) =& \frac{1}{2}\frac{\ln\det K_2(w^*)}{1-\alpha}\frac{1}{n} + o\bigg(\frac{1}{n}\bigg),\\
D_3(n) =& \frac{1}{2}\frac{\ln\det K_3(w^*)}{1-\alpha}\frac{1}{n} + o\bigg(\frac{1}{n}\bigg),
\end{align*}
where
\begin{align*}
K_2(w) =& \big(I_{xy}(w)-\alpha I_x(w)\big)\big(\alpha I_{xy}(w)+(1-2\alpha)I_x(w)\big)^{-1},\\
K_3(w) =& I_{xy}(w)\big(\alpha I_{xy}(w)+(1-\alpha)I_x(w)\big)^{-1}.
\end{align*}
\end{theorem}
The proof is in the appendix.

The theorem shows that, in all models, the dominant order is $1/n$,
which is the speed at which the error converges to zero.
The accuracy of Model 1 depends on the dimension of $\bar{w}$
instead of the position of $\bar{w}^*$;
note that, in the other models, the coefficients of the dominant order are functions of $w^*$.

Let us compare these error values.
We assume that Model 1 is based on the generative expression and uses the reduced parameter $\bar{w}$.
According to Theorem \ref{th:3asympError}, the asymptotic forms of the error functions are expressed as
\begin{align*}
D_i(n) = \frac{c_i}{n} +o\bigg(\frac{1}{n}\bigg),
\end{align*}
where $c_i$ is a positive constant.
When $c_i<c_j$, we define the magnitude relation between the error functions as
\begin{align*}
D_i(n)<D_j(n).
\end{align*}
The following theorem shows the relation among the three error functions;
\begin{theorem}
\label{th:comp3error}
If the nonzero eigenvalues of $I(w^*)I_x(w^*)^{-1}$ are all non negative,
the following inequality holds asymptotically:
\begin{align*}
D_3(n)<D_2(n)<D_1(n).
\end{align*}
\end{theorem}
The proof is in the appendix.
\section{Discussions}
\label{sec:dis}
\subsection{On the Magnitude Relation in Theorem \ref{th:comp3error}}
First, let us consider the magnitude relation in Theorem \ref{th:comp3error}.
A larger amount of training data obviously increases the accuracy of the estimation,
and a high dimensional parameter allows the model to be expressive and complex.
It is known that the asymptotic form of the error function depends on the number of data and the dimension of the parameter
in many cases,
and there is a trade-off between them.
For example, the generalization error $G(n)$ for the OV estimation is defined by
\begin{align*}
G(n) = E_n\bigg[ \int q(x)\ln \frac{q(x)}{p(x|X^n)}dx\bigg],
\end{align*}
where $X^n=\{x_1,\dots,x_n\}$, $E_n[\cdot]$ is the expectation over all training data $X^n$,
and $p(x|X^n)$ is the predictive distribution.
In the Bayes method, the predictive distribution is given by
\begin{align*}
p(x|X^n) =& \int p(x|w)p_o(w|X^n)dw,\\
p_o(w|X^n) =& \frac{\prod_{i=1}^np(x_i|w)\varphi(w|\eta)}{\int \prod_{i=1}^np(x_i|w)\varphi(w|\eta)dw}.
\end{align*}
Under the condition (A2), the asymptotic form of the generalization error is expressed as
\begin{align*}
G(n) = \frac{\dim w}{2n} + o\bigg(\frac{1}{n}\bigg),
\end{align*}
where $n$ is the number of data and $\dim w$ is the number of the parameter \citep{Schwarz,Rissanen,Clarke90,Levin}.
Obviously, the prediction is accurate when $n$ is large or $\dim w$ is small.

In the unsupervised LV estimation with the generative expression,
the task is to estimate all labels $Y^n=\{Y_1,Y_2\}=\{y_1,\dots,y_n\}$ based on $X^n=\{X_1,X_2\}$.
In the Bayes method, the estimated distribution of $Y^n$ is described by
\begin{align*}
p(Y^n|X^n) =& \int \prod_{i=1}^n\frac{p(x_i,y_i|w)}{p(x_i|w)}p_o(w|X^n)dw,
\end{align*}
and the error function is defined by
\begin{align*}
D_U(n) = \frac{1}{n}E_n\bigg[ \sum_{Y^n} q(Y^n|X^n)\ln \frac{q(Y^n|X^n)}{p(Y^n|X^n)}\bigg],
\end{align*}
where the true distribution of $Y^n$ is given by
\begin{align*}
q(Y^n|X^n) =& \prod_{i=1}^n \frac{q(x_i,y_i)}{q(x_i)}.
\end{align*}
Under the condition (A2), the error function has the following asymptotic form \citep{Yamazaki14a},
\begin{align*}
D_U(n) =& \frac{1}{2n}\ln \det I_{xy}(w^*)I_x(w^*)^{-1} +o\bigg(\frac{1}{n}\bigg).
\end{align*}
Since the rank of $I_{xy}(w^*)I_x(w^*)^{-1}$ is determined by the dimension of $w$,
the LV estimation is also accurate when $n$ is large or $\dim w$ is small.

Let us compare the three models from the perspective of the parameter dimension and the amount of data.
Since $\dim\bar{w} < \dim w$, Model 1 has an advantage in the parameter dimension.
On the other hand, the actual amount of data used for the estimation is larger in Models 2 and 3;
according to the second definition of $p(Y_2|D)$, Eq.~\ref{eq:2nddef} and Fig.~\ref{fig:rel_data_model}-(a),
the posterior distribution of Model 1 is constructed by only $\{X_1,Y_1\}$,
while those of Models 2 and 3 require $D=\{X_1,Y_1,X_2\}$.
Theorem \ref{th:comp3error} shows that, in order to improve accuracy, 
increasing the amount of data is more effective than reducing the dimension of the parameter.
Model 1 is thus at a disadvantage.

\subsection{The Effect of the Posterior Convergence on the Accuracy}
Next, we discuss convergence of the posterior distribution in Eq.~\ref{eq:posterior} and its effect on the accuracy.
The asymptotic form of the error function in Theorem \ref{th:3asympError} indicates the effect of the posterior.
According to its proof in Appendix,
the posterior converges to the Gaussian distribution $\mathcal{N}(\tilde{w},\tilde{\Sigma}/n)$ in law,
where $\tilde{w}$ is the maximum-likelihood estimator of $L(w,D)$ in Eq.~\ref{eq:posterior},
and $\tilde{\Sigma}$ is given by
\begin{align*}
\tilde{\Sigma}^{-1} =&
\begin{cases}
\alpha I_{y|x}(\bar{w}^*) & \text{Model 1}\\
\alpha I_{xy}(w^*)+(1-2\alpha)I_x(w^*) & \text{Model 2}\\
\alpha I_{xy}(w^*)+(1-\alpha)I_x(w^*) & \text{Model 3}
\end{cases}.
\end{align*}
The right-hand side corresponds to the inverse matrix in the coefficient.
For example,  $K_2(w^*)$ for Model 2 has the inverse matrix $\{\alpha I_{xy}(w^*)+(1-2\alpha)I_x(w^*)\}^{-1}$.
Therefore, the variance of the posterior, which shows the convergence speed, is one of the essential factor
to determine the accuracy.
In Model 1, the original form of the coefficient is $\ln \det I_{y|x}(\bar{w}^*)\{\alpha I_{y|x}(\bar{w}^*)\}^{-1}$,
and then $\dim \bar{w} \ln 1/\alpha$ appears instead of $\tilde{\Sigma}^{-1}$.

As shown in Eq.~\ref{eq:2nddef}, the posterior determines the difference of the models.
Then, the magnitude relation in Theorem \ref{th:comp3error} reflects the difference of the convergence speeds of the posterior distributions.
In order to visualize this difference,
let us experimentally compare the posterior distributions in the settings of Examples \ref{ex:data} and \ref{ex:model}.
The Markov chain Monte Carlo (MCMC) method was employed for obtaining parameter samples from the posterior.
The total number of data was $n=400$, and the ratio of labeled data was $\alpha=0.5$.
\begin{figure}[t]
\begin{minipage}[b]{.3\linewidth}
\centering
\includegraphics[width=\linewidth]{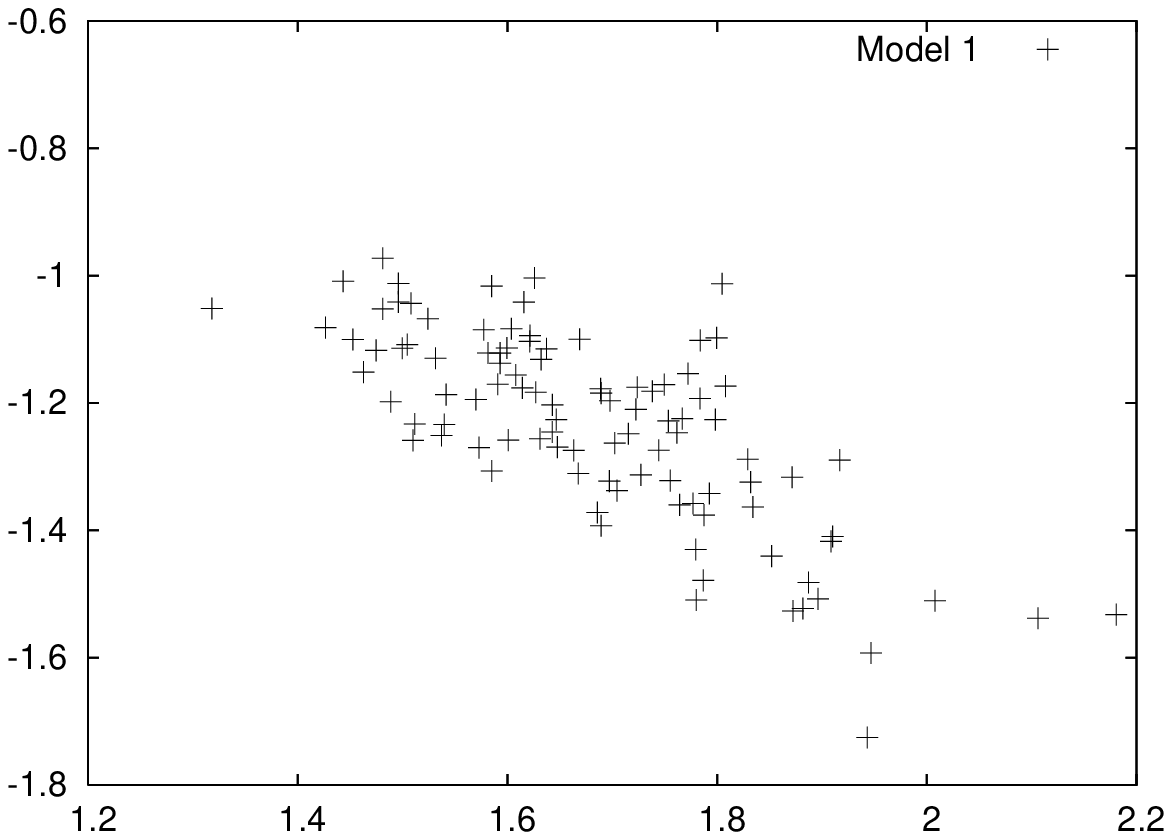}
\subcaption{Model 1}
\end{minipage}\;\;\;
\begin{minipage}[b]{.3\linewidth}
\centering
\includegraphics[width=\linewidth]{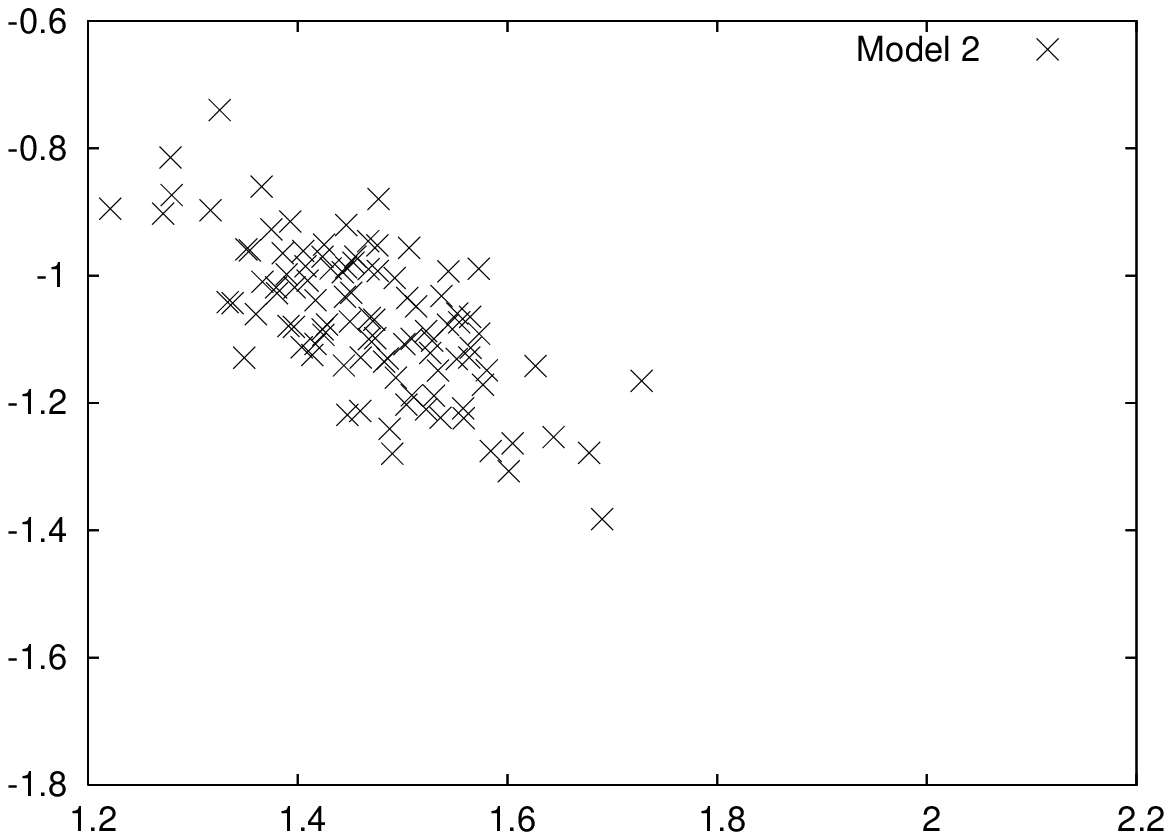}
\subcaption{Model 2}
\end{minipage}\;\;\;
\begin{minipage}[b]{.3\linewidth}
\centering
\includegraphics[width=\linewidth]{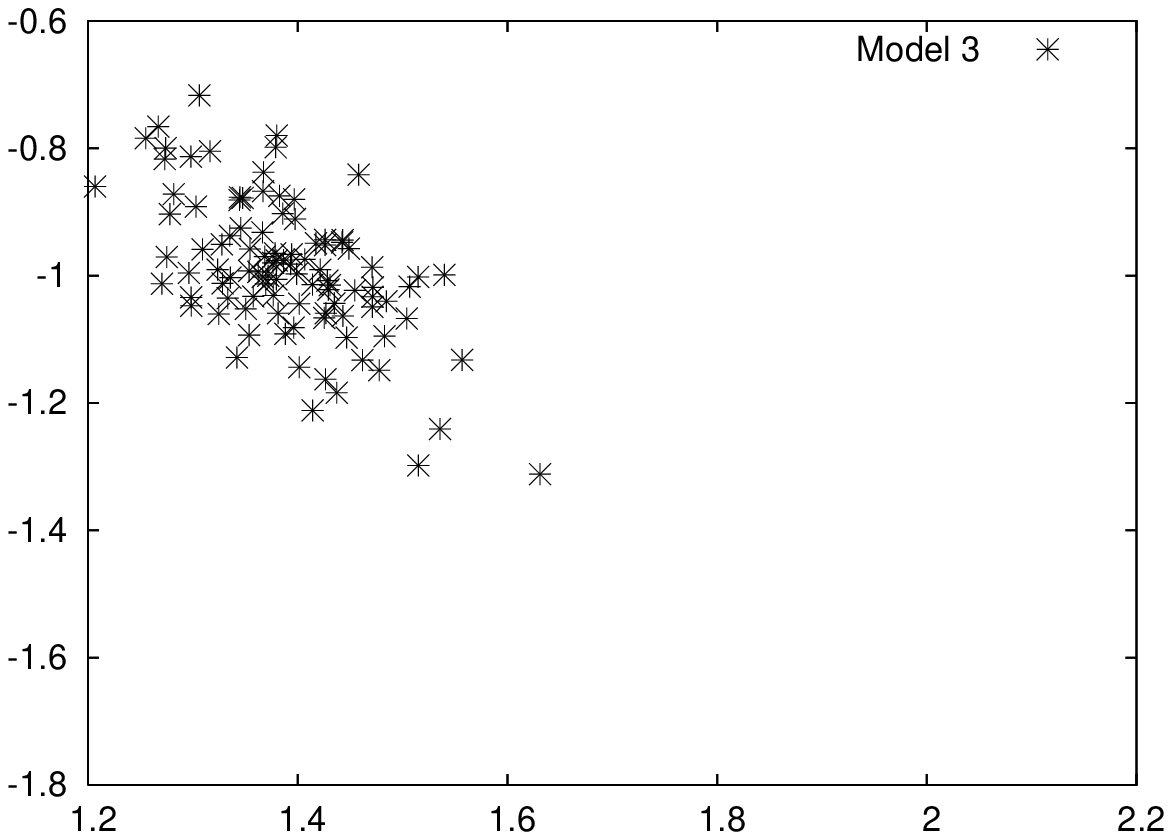}
\subcaption{Model 3}
\end{minipage}
\caption{Sampled points from the posterior distribution in the space of $\bar{w}$}
\label{fig:posterior}
\end{figure}
Fig.~\ref{fig:posterior} shows 100 sampled points from the posterior in each model.
To compare Model 1 with Models 2 and 3, we mapped samples of the parameter $w$
to the space of the reduced parameter $\bar{w}=(c_1,c_2)^\top$.
According to the calculation in Section \ref{sec:redundancy},
the dimension of $\bar{w}$ is $\dim \bar{w}=2$,
and the mapping is defined by
\begin{align*}
c_1 =& b_2 - b_1,\\
c_2 =& -\frac{1}{2}(b_2-b_1)(b_2+b_1) + \ln\frac{1-a_1}{a_1},
\end{align*}
where we assumed $\sigma=1$ for simplicity.
We conducted this MCMC sampling in 50 different data sets,
where each data set $X^n$ contains 200 labeled and 200 unlabeled data.
Let $\bar{\mu}$ be the empirical mean of the sampled points,
and $E[\bar{\mu}]$ be its expectation over 50 sets.
The true parameter was $\bar{w}^*=(1.5, -1.125)^\top$,
which was the convergence point of the posterior at $n\rightarrow\infty$.
\begin{table}[t]
\centering
\begin{tabular}{|c|c|c|}
\hline
models& $E[\bar{\mu}]$ & $E[(\bar{\mu}-w^*)^2]$ \\
\hline
Model 1 & $(1.55231, -1.16090)^\top$ & 0.12081\\
\hline
Model 2 & $(1.47616, -1.09729)^\top$ & 0.06214\\
\hline
Model 3 & $(1.48225, -1.09883)^\top$ & 0.05043\\
\hline
\end{tabular}
\caption{The convergence of the posterior distribution}
\label{tab:convergence}
\end{table}
Table \ref{tab:convergence} shows $E[\bar{\mu}]$ and $E[(\bar{\mu}-\bar{w}^*)^2]$.
The smaller $E[(\bar{\mu}-\bar{w}^*)^2]$ is, the faster the posterior converges.
Model 1 had the slowest convergence because its posterior was constructed without the unlabeled data
and the amount of the data was much smaller than that of the other models.
As Theorem \ref{th:comp3error} predicted, 
the convergence of Model 3 was faster than that of Model 2.
The posterior distributions of Models 2 and 3 were originally defined in the space of $w$.
Since the parameter $w$ was reducible to the lower dimensional $\bar{w}$, 
the discriminative expression for the labeled data $\prod_{i=1}^{\alpha n} p(y_i|x_i,w)$
had less information on the original parameter $w$ than the generative expression $\prod_{i=1}^{\alpha n} p(x_i,y_i|w)$.
This was the reason why Model 3 had better results than Model 2.
\subsection{Comparison with the Estimation without Labels}
In order to clarify the effect of the observable labels,
we compare the accuracy between Model 3 and the estimation of $Y_2$ without the labels $Y_1$.
The estimation is expressed as
\begin{align*}
p_{NL}(Y_2|X_1,X_2) =& \frac{\int \prod_{i=1}^{\alpha n} p(x_i|w)\prod_{i=\alpha n+1}^n p(x_i,y_i|w)\varphi(w|\eta)dw}
{\int \prod_{i=1}^n p(x_i|w)\varphi(w|\eta)dw},
\end{align*}
which is a generative expression.
The error function is then given by
\begin{align*}
D_{NL}(n) =& \frac{1}{(1-\alpha)n}E_D\bigg[ q(Y_2|D)\ln \frac{q(Y_2|D)}{p_{NL}(Y_2|X_1,X_2)}\bigg].
\end{align*}
This corresponds to the Type II estimation in \citet{Yamazaki14a},
and the asymptotic form of the error has been derived as
\begin{align*}
D_{NL}(n) =& \frac{1}{2}\frac{\ln\det K_4(w^*)}{1-\alpha}\frac{1}{n} + o\bigg(\frac{1}{n}\bigg)
\end{align*}
under the condition (A2), where
\begin{align*}
K_4(w) =& ((1-\alpha)I_{xy}(w)+\alpha I_x(w))I_x(w)^{-1}.
\end{align*}

The following lemma shows the quantitative difference between the estimations with and without the observable labels $Y_1$,
\begin{lemma}
\label{lem:compM3_NL}
Let assume that $I_{xy}(w^*)I_x(w^*)^{-1}$ has the eigen values $\lambda_1,\dots,\lambda_d$.
Under the same conditions of Theorem \ref{th:comp3error},
all eigen values are not less than one.
Then, the asymptotic error is described by
\begin{align}
D_{NL}(n) =& \frac{1}{2(1-\alpha)n}\sum_{i=1}^d \ln(\alpha + (1-\alpha)\lambda_i) + o\bigg(\frac{1}{n}\bigg), \label{eq:lambda_DNL}
\end{align}
and the magnitude relation to the error of Model 3 is given by
\begin{align*}
D_{NL}(n) > D_3(n).
\end{align*}
More precisely, the asymptotic difference of these error functions is described as
\begin{align}
D_{NL}(n) - D_3(n) =& \frac{1}{2(1-\alpha)n}\sum_{i=1}^d\ln\bigg\{ \alpha(1-\alpha)\bigg(\lambda_i+\frac{1}{\lambda_i}\bigg)+\alpha^2+(1-\alpha)^2\bigg\}\nonumber\\
&+ o\bigg(\frac{1}{n}\bigg),\label{eq:dif_M3_NL}
\end{align}
where the coefficient of the dominant term is positive
since the factor $\lambda_i+1/\lambda_i$ is the convex function with respect to $\lambda_i$
and has the minimum value at $\lambda_i=1$.
\end{lemma}
The proof is in the appendix.
Let us focus on the case, where the labels are informative
and the difference between the information matrices with and without labels is large.
The eigen value $\lambda_i$ increases from one since $\lambda_i=1$ for $I_{xy}(w^*)=I_x(w^*)$.
The asymptotic form in Eq.~\ref{eq:lambda_DNL} shows how the accuracy is adversely affected by this increase.
Therefore, $\lambda_i$ indicates the difficulty of the task in the unsupervised learning.
According to Eq.~\ref{eq:dif_M3_NL},
the difference of the error functions is also determined by the eigen values,
Because the factor $\lambda_i+1/\lambda_i$ is the increasing function for $\lambda_i\ge 1$,
the accuracy of the semi-supervised learning is significantly improved
when the task is difficult and $\lambda_i$ grows.
\section{Conclusion}
In semi-supervised learning, the given labels are used for the estimation of unobservable labels.
Depending on the expression of the labeled data in the likelihood function,
we have three approaches: generative, discriminative, and their hybrid models.
In the present paper, we focus on the Bayes method to estimate the labels in a distribution-based manner,
and derive the asymptotic form of the error function measuring the accuracy with the Kullback-Leibler divergence.
Comparing the asymptotic forms in the three models,
we prove that the generative model performs better when the model is well specified.
The asymptotic error depends on the amount of data and the dimension of the parameter,
and there is a trade-off between them.
The discriminative model does not require high dimensional parameter due to its simple expression
while the generative model uses more data for the estimation.
The magnitude relation theoretically indicates that increasing the amount of data is more effective
than reducing the dimension of the parameter in order to improve accuracy.
%
%
\section*{Appendix}
This section shows the proofs of the theorems.
\subsection*{Proof of Theorem \ref{th:3asympError}}
First, we derive the asymptotic form of $D_1(n)$.
Define the free-energy function by
\begin{align*}
F_{y|x}(n) =& E_n\bigg[ \ln\prod_{i=1}^n q(y_i|x_i) -\ln\int\prod_{i=1}^n p(y_i|x_i,\bar{w})\varphi(\bar{w}|\eta)d\bar{w} \bigg],
\end{align*}
where the expectation is
\begin{align*}
E_n[f(X^n,Y^n)] =& \int \sum_{Y^n} f(X^n,Y^n) q(Y^n|X^n)q(X^n)dX^n.
\end{align*}
The error function $D_1(n)$ can be rewritten as
\begin{align}
(1-\alpha)n D_1(n) =& E_D\bigg[\sum_{Y_2} q(Y_2|D)\bigg\{
\ln\frac{\prod_{i=1}^n q(y_i|x_i)}
{\int \prod_{i=1}^n p(y_i|x_i,\bar{w})\varphi(\bar{w}|\eta)d\bar{w}}\nonumber\\
&-\ln\frac{\prod_{i=1}^{\alpha n} q(y_i|x_i)}
{\int \prod_{i=1}^{\alpha n} p(y_i|x_i,\bar{w})\varphi(\bar{w}|\eta)d\bar{w}}\}\bigg\}\bigg]\nonumber\\
=& F_{y|x}(n) - F_{y|x}(\alpha n). \label{eq:DFF}
\end{align}
It is sufficient to calculate the asymptotic form of $F_{y|x}(n)$.

The maximum-likelihood estimator is defined by
\begin{align*}
\hat{w}_n =& \arg\max_{\bar{w}} \prod_{i=1}^n p(y_i|x_i,\bar{w}).
\end{align*}
Due to (A1), the estimator converges to $\bar{w}^*$,
which means that the essential parameter area for the integration
is the neighborhood of $\bar{w}^*$ and $\hat{w}_n$.
According to the Taylor expansion at $w=\hat{w}_n$,
\begin{align*}
E_n&\bigg[ \ln \int \prod_{i=1}^n p(y_i|x_i,\bar{w})\varphi(\bar{w}|\eta)d\bar{w}\bigg]\\
=& E_n\bigg[\ln\int\exp\bigg\{n\frac{1}{n}\sum_{i=1}^n \ln p(y_i|x_i,\bar{w})\bigg\}\varphi(\bar{w}|\eta)d\bar{w}\bigg]\\
=& E_n\bigg[\ln\int\exp\bigg\{n\frac{1}{n}\sum_{i=1}^n \ln p(y_i|x_i,\hat{w}_n)\\
&+n\frac{1}{n}(\bar{w}-\hat{w}_n)^\top\frac{\partial}{\partial \bar{w}}\sum_{i=1}^n \ln p(y_i|x_i,\hat{w}_n)\\
&+n\frac{1}{2}(\bar{w}-\hat{w}_n)^\top\frac{1}{n}\frac{\partial^2}{\partial \bar{w}^2}
\sum_{i=1}^n \ln p(y_i|x_i,\hat{w}_n)(w-\hat{w}_n) +r_1(\bar{w}) \bigg\}\varphi(\bar{w}|\eta)d\bar{w}\bigg],
\end{align*}
where $r_1(\bar{w})$ is the remainder term.
Based on the saddle-point approximation,
\begin{align*}
E_n&\bigg[ \ln \int \prod_{i=1}^n p(y_i|x_i,\bar{w})\varphi(\bar{w}|\eta)d\bar{w}\bigg]\\
=& E_n\bigg[\sum_{i=1}^n \ln p(y_i|x_i,\hat{w}_n)\bigg]\\
&+ E_n\bigg[\ln\int\exp(nr_1(\bar{w}))\varphi(\bar{w}|\eta)\mathcal{N}(\hat{w}_n,(nI_{y|x}(\bar{w}^*))^{-1})d\bar{w}\bigg]\\
&- \bigg\{\frac{\dim \bar{w}}{2}\ln n -\frac{\dim \bar{w}}{2}\ln 2\pi +\frac{1}{2}\ln\det I_{y|x}(\bar{w}^*)\bigg\} +o(1)\\
=& E_n\bigg[\sum_{i=1}^n \ln p(y_i|x_i,\hat{w}_n)\bigg]
-\frac{\dim \bar{w}}{2}\ln n \\
&+\frac{\dim \bar{w}}{2}\ln 2\pi -\frac{1}{2}\ln\det I_{y|x}(\bar{w}^*) +\ln\varphi(\bar{w}^*) +o(1),
\end{align*}
where $\mathcal{N}(\mu,\Sigma)$ is a $\dim \bar{w}$-dimensional Gaussian distribution
with mean $\mu\in R^{\dim \bar{w}}$ and variance-covariance matrix $\Sigma$.
Thus, we obtain
\begin{align*}
F_{y|x}(n) =& E_n\bigg[ \sum_{i=1}^n \ln \frac{q(y_i|x_i)}{p(y_i|x_i,\hat{w}_n)}\bigg]\\
&+ \frac{\dim \bar{w}}{2}\ln n -\frac{\dim \bar{w}}{2}\ln 2\pi +\frac{1}{2}\ln\det I_{y|x}(\bar{w}^*) -\ln \varphi(\bar{w}^*|\eta) + o(1).
\end{align*}
According to the Taylor expansion at $\hat{w}_n$,
\begin{align*}
E_n&\bigg[ \sum_{i=1}^n \ln p(y_i|x_i,\bar{w}^*) \bigg]\\
=& E_n\bigg[ \sum_{i=1}^n \ln p(y_i|x_i,\hat{w}_n)\\
& + (\bar{w}^*-\hat{w}_n)^\top \frac{\partial}{\partial \bar{w}}\sum_{i=1}^n \ln p(y_i|x_i,\hat{w}_n)\\
& + \frac{1}{2} (\bar{w}^*-\hat{w}_n)^\top \frac{\partial^2}{\partial \bar{w}^2}\sum_{i=1}^n \ln p(y_i|x_i,\hat{w}_n) (\bar{w}^*-\hat{w}_n)
+ r_2(\bar{w})\bigg]\\
=& E_n\bigg[ \sum_{i=1}^n \ln p(y_i|x_i,\hat{w}_n)\bigg]
- \frac{n}{2}E_n\bigg[ (\bar{w}^*-\hat{w}_n)^\top I_{y|x}(\bar{w}^*) (\bar{w}^*-\hat{w}_n)\bigg] + o(1),
\end{align*}
where $r_2(\bar{w})$ is the remainder term.
The estimator $\hat{w}_n$ has asymptotic normality,
and it converges to the Gaussian distribution with mean $w_0$
and variance--covariance matrix $(nI_{y|x}(\bar{w}^*)^{-1})^{-1}$.
It holds that
\begin{align*}
E_n\bigg[ \sum_{i=1}^n \ln p(y_i|x_i,\bar{w}^*) \bigg]
= E_n\bigg[ \sum_{i=1}^n \ln p(y_i|x_i,\hat{w}_n)\bigg] - \frac{\dim \bar{w}}{2} + o(1).
\end{align*}
The free-energy function has the asymptotic form
\begin{align*}
F_{y|x}(n) =& \frac{\dim \bar{w}}{2}\ln n -\frac{\dim \bar{w}}{2}\ln 2\pi e +\frac{1}{2}\ln\det I_{y|x}(\bar{w}^*) -\ln \varphi(\bar{w}^*|\eta) + o(1),
\end{align*}
which is consistent with the form derived in \citep{Clarke90}.
Based on the relation in Eq.~\ref{eq:DFF}, we obtain
\begin{align*}
(1-\alpha)n D_1(n) =& - \frac{\dim \bar{w}}{2}\ln \alpha + o(1),
\end{align*}
which is the asymptotic form of $D_1(n)$.

In the similar way, we derive the asymptotic forms of $D_2(n)$ and $D_3(n)$.
Define the free-energy functions as
\begin{align*}
F_{xy}(n) =& E_n\bigg[ \ln\prod_{i=1}^n q(x_i,y_i) -\ln\int\prod_{i=1}^n p(x_i,y_i|w)\varphi(w|\eta)dw\bigg],\\
F_{xy,x}(n) =& E_n\bigg[ \ln \prod_{i=1}^{\alpha n} q(x_i,y_i) \prod_{i=\alpha n+1}^n q(x_i)\\
&-\ln\int\prod_{i=1}^{\alpha n}p(x_i,y_i|w)\prod_{i=\alpha n+1}^n p(x_i|w)\varphi(w|\eta)dw\bigg].
\end{align*}
The error function $D_3(n)$ is rewritten as
\begin{align}
(1-\alpha)nD_3(n) =& E_D\bigg[\sum_{Y_2} q(Y_2|D)\bigg\{
\ln\frac{\prod_{i=1}^n q(x_i,y_i)}
{\int \prod_{i=1}^n p(x_i,y_i|w)\varphi(w|\eta)dw}\nonumber\\
&-\ln\frac{\prod_{i=1}^{\alpha n}q(x_i,y_i)\prod_{i=\alpha n+1}^n q(x_i)}
{\int \prod_{i=1}^{\alpha n}p(x_i,y_i|w)\prod_{i=\alpha n+1}^n p(x_i|w)\varphi(w|\eta)dw}\bigg\}\bigg]\nonumber\\
=& F_{xy}(n) - F_{xy,x}(n).\label{eq:DFF3}
\end{align}
The maximum-likelihood estimator is defined by
\begin{align*}
\hat{w}_{xy} =& \arg\max_w \prod_{i=1}^n p(x_i,y_i|w).
\end{align*}
Due to (A2), the estimators $\hat{w}_{xy}$ and $\hat{w}_3$ converge to $w^*$,
which means that the essential parameter area for the integration
is the neighborhood of $w^*$, $\hat{w}_{xy}$, and $\hat{w}_3$.
According to the Taylor expansion at $w=\hat{w}_{xy}$,
\begin{align*}
E_n&\bigg[ \ln \int \prod_{i=1}^n p(x_i,y_i|w)\varphi(w|\eta)dw\bigg]\\
=& E_n\bigg[\ln\int\exp\bigg\{n\frac{1}{n}\sum_{i=1}^n \ln p(x_i,y_i|w)\bigg\}\varphi(w|\eta)dw\bigg]\\
=& E_n\bigg[\ln\int\exp\bigg\{n\frac{1}{n}\sum_{i=1}^n \ln p(x_i,y_i|\hat{w}_{xy})\\
&+n\frac{1}{n}(w-\hat{w}_{xy})^\top\frac{\partial}{\partial w}\sum_{i=1}^n \ln p(x_i,y_i|\hat{w}_{xy})\\
&+n\frac{1}{2}(w-\hat{w}_{xy})^\top\frac{1}{n}\frac{\partial^2}{\partial w^2}
\sum_{i=1}^n \ln p(x_i,y_i|\hat{w}_{xy})(w-\hat{w}_{xy}) +r_1(w) \bigg\}\varphi(w|\eta)dw\bigg],
\end{align*}
where $r_1(w)$ is the remainder term.
Based on the saddle-point approximation,
\begin{align*}
E_n&\bigg[ \ln \int \prod_{i=1}^n p(x_i,y_i|w)\varphi(w|\eta)dw\bigg]\\
=& E_n\bigg[\sum_{i=1}^n \ln p(x_i,y_i|\hat{w}_{xy})\bigg]\\
&+ E_n\bigg[\ln\int\exp(nr_1(w))\varphi(w|\eta)\mathcal{N}(\hat{w}_{xy},(nI_{xy}(w^*))^{-1})dw\bigg]\\
&- \bigg\{\frac{\dim w}{2}\ln n -\frac{\dim w}{2}\ln 2\pi +\frac{1}{2}\ln\det I_{xy}(w^*)\bigg\} +o(1)\\
=& E_n\bigg[\sum_{i=1}^n \ln p(x_i,y_i|\hat{w}_{xy})\bigg]
-\frac{\dim w}{2}\ln n \\
&+\frac{\dim w}{2}\ln 2\pi -\frac{1}{2}\ln\det I_{xy}(w^*) +\ln\varphi(w^*) +o(1),
\end{align*}
where $\mathcal{N}(\mu,\Sigma)$ is a $\dim w$-dimensional Gaussian distribution
with mean $\mu\in R^{\dim w}$ and variance-covariance matrix $\Sigma$.
Then, we obtain
\begin{align*}
F_{xy}(n) =& E_n\bigg[ \sum_{i=1}^n \ln\frac{p(x_i,y_i|w^*)}{p(x_i,y_i|\hat{w}_{xy})}\bigg]\\
&+ \frac{\dim w}{2}\ln n -\frac{\dim w}{2}\ln 2\pi +\frac{1}{2}\ln\det I_{xy}(w^*) -\ln \varphi(w^*|\eta) + o(1).
\end{align*}
According to the Taylor expansion at $\hat{w}_{xy}$,
\begin{align*}
E_n&\bigg[ \sum_{i=1}^n \ln p(x_i,y_i|w^*) \bigg]\\
=& E_n\bigg[ \sum_{i=1}^n \ln p(x_i,y_i|\hat{w}_{xy})\\
& + (w^*-\hat{w}_{xy})^\top \frac{\partial}{\partial w}\sum_{i=1}^n \ln p(x_i,.y_i|\hat{w}_{xy})\\
& + \frac{1}{2} (w^*-\hat{w}_{xy})^\top \frac{\partial^2}{\partial w^2}\sum_{i=1}^n \ln p(x_i,y_i|\hat{w}_{xy}) (w^*-\hat{w}_{xy})
+ r_2(w)\bigg]\\
=& E_n\bigg[ \sum_{i=1}^n \ln p(x_i,y_i|\hat{w}_{xy})\bigg]
- \frac{n}{2}E_n\bigg[ (w^*-\hat{w}_{xy})^\top I_{xy}(w^*) (w^*-\hat{w}_{xy})\bigg] + o(1),
\end{align*}
where $r_2(w)$ is the remainder term.
The estimator $\hat{w}_{xy}$ has asymptotic normality,
and it converges to the Gaussian distribution with mean $w^*$
and variance--covariance matrix $(nI_{xy}(w^*))^{-1}$.
It holds that
\begin{align*}
E_n\bigg[ \sum_{i=1}^n \ln p(x_i,y_i|w^*) \bigg]
=& E_n\bigg[ \sum_{i=1}^n \ln p(x_i,y_i|\hat{w}_{xy})\bigg] - \frac{\dim w}{2} + o(1).
\end{align*}
The free-energy function has the asymptotic form
\begin{align*}
F_{xy}(n) =& - \frac{\dim w}{2} + \frac{\dim w}{2}\ln n -\frac{\dim w}{2}\ln 2\pi\\
& +\frac{1}{2}\ln\det I_{xy}(w^*) -\ln \varphi(w^*|\eta) + o(1).
\end{align*}
In the same way, we obtain that
\begin{align*}
F_{xy,x}(n)=& - \frac{\dim w}{2} + \frac{\dim w}{2}\ln n -\frac{\dim w}{2}\ln 2\pi\\
& +\frac{1}{2}\ln\det \{\alpha I_{xy}(w^*)+(1-\alpha)I_x(w^*)\} -\ln \varphi(w^*|\eta) + o(1).
\end{align*}
Based on the relation in Eq.~\ref{eq:DFF3}, we obtain that
\begin{align*}
(1-\alpha)nD_3(n) =& \frac{1}{2}\ln\det\bigg\{I_{xy}(w^*)\big(\alpha I_{xy}(w^*)+(1-\alpha)I_x(w^*)\big)^{-1}\bigg\} +o(1),
\end{align*}
which is the asymptotic form of $D_3(n)$.

Define the free-energy functions by
\begin{align*}
F_{y|x,xy}(n) =& E_n\bigg[ \ln\prod_{i=1}^{\alpha n} q(y_i|x_i)\prod_{i=\alpha n+1}^n q(x_i,y_i)\\
& -\ln\int\prod_{i=1}^{\alpha n} p(y_i|x_i,w)\prod_{i=\alpha n+1}^n p(x_i,y_i|w)\varphi(w|\eta)dw \bigg],\\
F_{y|x,x}(n) =& E_n\bigg[ \ln\prod_{i=1}^{\alpha n} q(y_i|x_i)\prod_{i=\alpha n+1}^n q(x_i)\\
& -\ln\int\prod_{i=1}^{\alpha n} p(y_i|x_i,w)\prod_{i=\alpha n+1}^n p(x_i|w)\varphi(w|\eta)dw \bigg].
\end{align*}
The error function can be rewritten as
\begin{align}
(1-\alpha)nD_2(n) =& F_{y|x,xy}(n)-F_{y|x,x}(n). \label{eq:DFF2}
\end{align}
The maximum-likelihood estimator is defined by
\begin{align*}
\hat{w}_{y|x,xy} =& \arg\max_w \prod_{i=1}^{\alpha n}p(y_i|x_i,w)\prod_{i=\alpha n+1}^n p(x_i,y_i|w).
\end{align*}
Due to (A2), the estimators $\hat{w}_{y|x,xy}$ and $\hat{w}_2$ converge to $w^*$,
which means that the essential parameter area for the integration
is the neighborhood of $w^*$, $\hat{w}_{y|x,xy}$, and $\hat{w}_2$.
According to the Taylor expansion and the saddle-point approximation,
the free-energy functions have the following asymptotic forms:
\begin{align*}
F_{y|x,xy}(n)=& - \frac{\dim w}{2} + \frac{\dim w}{2}\ln n -\frac{\dim w}{2}\ln 2\pi\\
& +\frac{1}{2}\ln\det \{I_{xy}(w^*)-\alpha I_x(w^*)\} -\ln \varphi(w^*|\eta) + o(1),\\
F_{y|x,xy}(n)=& - \frac{\dim w}{2} + \frac{\dim w}{2}\ln n -\frac{\dim w}{2}\ln 2\pi\\
& +\frac{1}{2}\ln\det \{\alpha I_{xy}(w^*)+(1-2\alpha) I_x(w^*)\} -\ln \varphi(w^*|\eta) + o(1).
\end{align*}
Based on the relation in Eq.~\ref{eq:DFF2}, we obtain that
\begin{align*}
(1-\alpha)nD_2(n) =& \frac{1}{2}\ln\det \bigg\{K_{21}(w^*)K_{22}(w^*)^{-1}\bigg\} +o(1),\\
K_{21}(w) =& I_{xy}(w) -\alpha I_x(w),\\
K_{22}(w) =& \alpha I_{xy}(w)+(1-2\alpha)I_x(w),
\end{align*}
which is the asymptotic form of $D_3(n)$.
{\bf (End of Proof)}
\subsection*{Proof of Theorem \ref{th:comp3error}}
According to the condition,
let the eigenvalues of $I(w^*)I_x(w^*)^{-1}$ be
\begin{align*}
\sigma_1 \ge \sigma_2 \ge \dots \ge \sigma_{\bar{d}}>0,\\
\sigma_{\bar{d}+1}=\dots=\sigma_d=0,
\end{align*}
where $\bar{d}=\dim \bar{w}$.
First, we compare $D_1(n)$ and $D_3(n)$.
Focusing on the factor $\ln\det K_3(w^*)$ of the dominant term in $D_3(n)$,
we obtain
\begin{align*}
\ln \det K_3(w^*) =& \ln\det I_{xy}(w^*) - \ln \det \big\{\alpha I_{xy}(w^*)+(1-\alpha) I_x(w^*)\big\}\\
=& \ln\det\big\{I(w^*)+I_x(w^*)\big\} -\ln\det\big\{ \alpha I(w^*) +I_x(w^*) \big\}\\
=& \ln\det\big\{ I(w^*)I_x(w^*)^{-1} +E\big\} -\ln\det\big\{\alpha I(w^*)I_x(w^*)^{-1} + E\big\}\\
=& \sum_{i=1}^d\ln(\sigma_i+1)-\sum_{i=1}^d \ln(\alpha \sigma_i+1)\\
=& \sum_{i=1}^{\bar{d}} \ln\frac{\sigma_i+1}{\alpha \sigma_i +1}\\
=& \bar{d}\ln\frac{1}{\alpha} + \sum_{i=1}^{\bar{d}}\ln\bigg( 1+\frac{1-1/\alpha}{\sigma_i+1/\alpha}\bigg),
\end{align*}
where $E$ is the $d\times d$ unit matrix and the relation $I(w)=I_{xy}(w)-I_x(w)$ was applied.
Because $\frac{1-1/\alpha}{\sigma_i+1/\alpha}<0$, the second term in the last expression is less than zero.
Thus,
\begin{align*}
\ln \det K_3(w^*) < \bar{d}\ln 1/\alpha,
\end{align*}
which shows that
\begin{align*}
D_3(n) - D_1(n) =& \frac{1}{2}\big\{\ln \det K_3(w^*)-\bar{d}\ln 1/\alpha\big\}\frac{1}{(1-\alpha)n} + o\bigg(\frac{1}{n}\bigg)<0.
\end{align*}
Next, we compare $D_1(n)$ and $D_2(n)$.
\begin{align*}
\ln \det K_2(w) =& \ln\det\big\{I_{xy}(w^*)-\alpha I_x(w^*)\big\}-\ln\det\big\{\alpha I_{xy}(w^*)+(1-2\alpha)I_x(w^*)\big\}\\
=& \ln\det\big\{ I(w^*)+(1-\alpha)I_x(w^*)\big\}-\ln\det\big\{\alpha I(w^*)+(1-\alpha)I_x(w^*)\big\}\\
=& \ln\det\big\{ I(w^*)I_x(w^*)^{-1}+(1-\alpha)E\big\}\\
& -\ln\det\big\{ \alpha I(w^*)I_x(w^*)^{-1}+(1-\alpha)E\big\}\\
=& \sum_{i=1}^d \ln(\sigma_i+(1-\alpha))-\sum_{i=1}^d\ln(\alpha\sigma_i+(1-\alpha))\\
=& \bar{d}\ln \frac{1}{\alpha} + \sum_{i=1}^{\bar{d}}\ln\frac{\sigma_i+1-\alpha}{\sigma_i+(1-\alpha)/\alpha}\\
=& \bar{d}\ln \frac{1}{\alpha} + \sum_{i=1}^{\bar{d}}\ln\bigg(1+\frac{(1-\alpha)(1-1/\alpha)}{\sigma_i+(1-\alpha)/\alpha}\bigg)\\
=& \bar{d}\ln \frac{1}{\alpha} + \sum_{i=1}^{\bar{d}}\ln\bigg(1+\frac{1-1/\alpha}{\sigma_i/(1-\alpha)+1/\alpha}\bigg).
\end{align*}
Because $\frac{1-1/\alpha}{\sigma_i/(1-\alpha)+1/\alpha}<0$,
the second term in the last expression is less than zero.
Thus,
\begin{align*}
\ln \det K_2(w^*) < \bar{d}\ln 1/\alpha,
\end{align*}
which shows that
\begin{align*}
D_2(n) - D_1(n)  =& \frac{1}{2}\big\{\ln \det K_2(w)-\bar{d}\ln 1/\alpha\big\}\frac{1}{(1-\alpha)n} +o\bigg(\frac{1}{n}\bigg)<0.
\end{align*}
Comparing $K_3(w^*)$ and $K_2(w^*)$,
we find that
\begin{align*}
\sum_{i=1}^{\bar{d}}\ln\bigg( 1+\frac{1-1/\alpha}{\sigma_i+1/\alpha}\bigg)
< \sum_{i=1}^{\bar{d}}\ln\bigg(1+\frac{1-1/\alpha}{\sigma_i/(1-\alpha)+1/\alpha}\bigg).
\end{align*}
Therefore, $\ln \det K_3(w^*)<\ln \det K_2(w^*)$, which shows that $D_3(n)<D_2(n)$.
{\bf (End of Proof)}
\subsection*{Proof of Lemma \ref{lem:compM3_NL}}
Focusing on the factor $\ln \det K_4(w^*)$ of the dominant term in $D_{NL}(n)$,
we obtain
\begin{align*}
\ln \det K_4(w^*) =& \ln \det \big( \alpha E + (1-\alpha)I_{xy}(w^*)I_x(w^*)^{-1}\big)\\
=& \sum_{i=1}^d \ln (\alpha +(1-\alpha)\lambda_i),
\end{align*}
which proves Eq.\ref{eq:lambda_DNL}.
Using the eigen values $\lambda_i$,
we rewrite the factor of the dominant term $\ln \det K_3(w^*)$ as
\begin{align*}
\ln \det K_3(w^*) =& I_{xy}(w^*)(\alpha I_{xy}(w^*) + (1-\alpha)I_x(w^*))^{-1}\\
=& -\ln \det (\alpha E +(1-\alpha)I_x(w^*)I_{xy}(w^*)^{-1})\\
=& -\sum_{i=1}^d \ln \bigg(\alpha + (1-\alpha)\frac{1}{\lambda_i}\bigg).
\end{align*}
The difference of the coefficients in $D_{NL}(n)-D_3(n)$ is expressed as
\begin{align*}
\ln \det K_4(w^*)-\ln \det K_3(w^*) =& \sum_{i=1}^d \bigg\{ \ln(\alpha+(1-\alpha)\lambda_i)+\ln\bigg(\alpha+(1-\alpha)\frac{1}{\lambda_i}\bigg)\bigg\}\\
=& \sum_{i=1}^d \ln \bigg(\alpha+(1-\alpha)\lambda_i\bigg)\bigg(\alpha+(1-\alpha)\frac{1}{\lambda_i}\bigg)\\
=& \sum_{i=1}^d \ln \bigg\{ \alpha(1-\alpha)\bigg(\lambda_i+\frac{1}{\lambda_i}\bigg)+\alpha^2+(1-\alpha)^2\bigg\},
\end{align*}
which shows the difference in Eq.~\ref{eq:dif_M3_NL} and $D_{NL}(n)>D_3(n)$.
{\bf (End of Proof)}
\bibliography{LearningTheory}   
\bibliographystyle{natbib}
\end{document}